\documentclass[10pt,twocolumn,letterpaper]{article}

\usepackage[pagenumbers]{iccv} 

\usepackage{listings}
\usepackage{xcolor}

\usepackage{multirow}

\definecolor{iccvblue}{rgb}{0.21,0.49,0.74}
\usepackage[pagebackref,breaklinks,colorlinks,allcolors=iccvblue]{hyperref}

\def\paperID{9085} 
\def\confName{ICCV}
\def\confYear{2025}

\title{Beyond Audio and Pose:\\ A General-Purpose Framework for Video Synchronization}

\author{
Yosub Shin \quad Igor Molybog \\
University of Hawai'i at Manoa \\
{\tt\small \{yosubs, molybog\}@hawaii.edu}
}

\begin{document}
\maketitle
\begin{abstract}
Video synchronization—aligning multiple video streams capturing the same event from different angles—is crucial for applications such as reality TV show production, sports analysis, surveillance, and autonomous systems. Prior work has heavily relied on audio cues or specific visual events, limiting applicability in diverse settings where such signals may be unreliable or absent. Additionally, existing benchmarks for video synchronization lack generality and reproducibility, restricting progress in the field. In this work, we introduce \textbf{VideoSync}, a video synchronization framework that operates independently of specific feature extraction methods, such as human pose estimation, enabling broader applicability across different content types. We evaluate our system on newly composed datasets covering single-human, multi-human, and non-human scenarios, providing both the methodology and code for dataset creation to establish reproducible benchmarks. Our analysis reveals biases in prior SOTA work, particularly in SeSyn-Net's preprocessing pipeline, leading to inflated performance claims. We correct these biases and propose a more rigorous evaluation framework, demonstrating that \textbf{VideoSync} outperforms existing approaches, including SeSyn-Net, under fair experimental conditions. Additionally, we explore various synchronization offset prediction methods, identifying a convolutional neural network (CNN)-based model as the most effective. Our findings advance video synchronization beyond domain-specific constraints, making it more generalizable and robust for real-world applications.
\end{abstract}    
\section{Introduction}\label{sec:intro}

Video synchronization—the task of aligning multiple video recordings of the same event—plays a critical role in multi-camera applications, including sports analytics, surveillance, and autonomous systems. One particularly demanding use case arises in reality TV production, where editors must synchronize footage from 50 to 100 cameras along with 20 to 50 audio channels, often spanning many hours. This process must be completed under intense time constraints, and the accuracy of synchronization is critical down to the frame level to ensure seamless editing and storytelling. Traditional methods rely on audio cues or explicit visual markers such as flashing lights, but these approaches break down when such signals are unavailable or unreliable—common in silent footage, underwater recordings, or multi-view autonomous driving scenarios. While recent methods attempt to use visual features, they often depend on specific hand-crafted embeddings like human pose estimations, limiting their applicability to broader contexts.

Despite its importance, video synchronization research lacks standardized benchmarks. Prior benchmarks, such as those used in SeSyn-Net, neither provide publicly available code nor cover a diverse range of synchronization scenarios. Additionally, we identify methodological flaws in SeSyn-Net's preprocessing pipeline that lead to inflated performance claims. Specifically, its preprocessing introduces biases that artificially improve synchronization accuracy. We correct these biases and provide an improved evaluation framework to ensure fair comparisons.

To advance the field, we propose \textbf{VideoSync}, a general-purpose video synchronization method that does not rely on domain-specific feature extraction techniques. Instead, we evaluate multiple self-supervised video embedding models—CARL~\citep{SCL} and InternVideo2~\citep{internvideo2}—to determine their effectiveness in synchronization tasks. We introduce new datasets that include single-human, multi-human, and non-human video pairs, addressing the limitations of existing benchmarks. Furthermore, we explore various synchronization offset prediction strategies, finding that a CNN-based approach consistently outperforms other methods.

\subsection{Contributions}
Our work makes the following key contributions:
\begin{itemize}
    \item \textbf{New Benchmark Datasets \& Reproducibility Kit:}
    We compose diverse datasets covering single-human, multi-human, and non-human scenarios. We provide both the methodology and code for dataset creation, establishing reproducible benchmarks for future research.
    \item \textbf{Evaluation of Prior Work:}
    We identify and correct biases in SeSyn-Net's preprocessing pipeline, demonstrating that its reported performance is inflated. We propose a fairer experimental setup and reassess its effectiveness.
    \item \textbf{General-Purpose Synchronization Approach:}
    \textbf{VideoSync} does not rely on task-specific features like human pose extraction, enabling synchronization in a broader range of videos. We evaluate multiple video embedding models, demonstrating their strengths and weaknesses.
    \item \textbf{Effective Synchronization Offset Prediction:}
    We experiment with various approaches for predicting synchronization offsets and identify a CNN-based method that consistently outperforms prior methods, including SeSyn-Net.
\end{itemize}

By addressing these challenges, our work establishes a more robust and generalizable foundation for video synchronization research.

\section{Overview of the Proposed Approach}
\begin{figure*}
    \centering
    \includegraphics[width=1.00\linewidth]{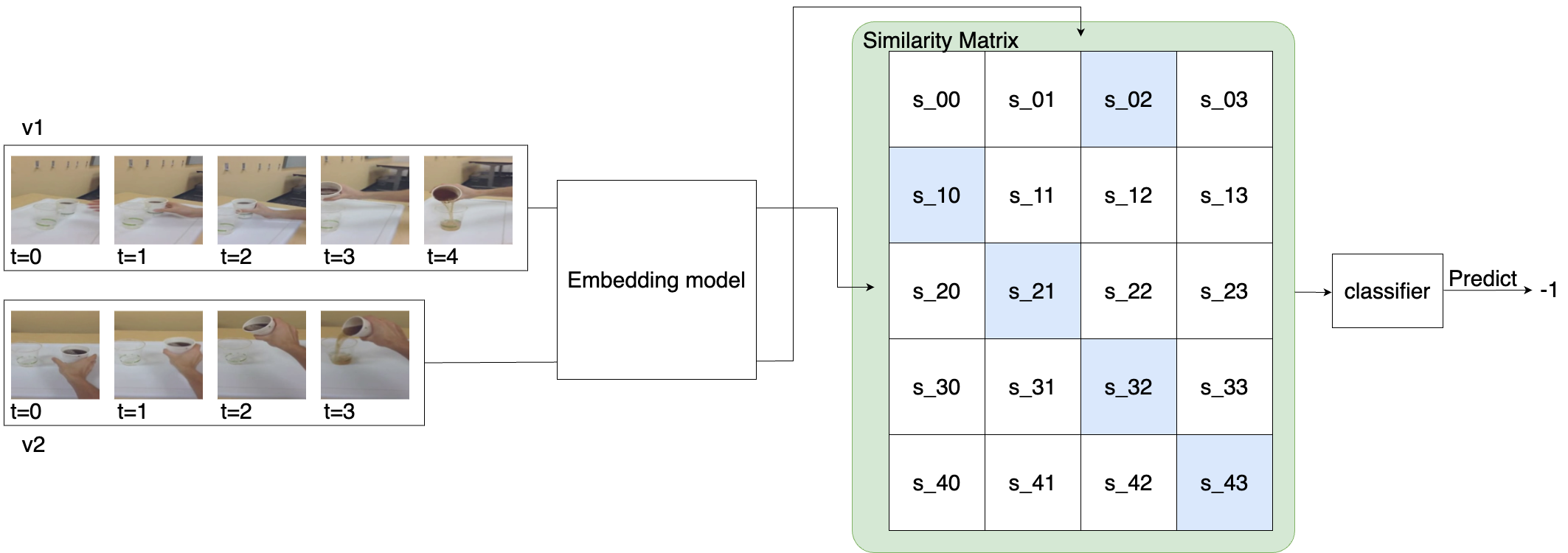}
    \caption{Overview of our video synchronization pipeline, \textbf{VideoSync}. The embedding model outputs embedding vectors at the frame level. Each entry in the similarity matrix represents the negative Euclidean distance between a frame from video $v_1$ and a frame from video $v_2$. For example, $s_{42}$ corresponds to the negative distance between the 4th frame of $v_1$ and the 2nd frame of $v_2$. The softmaxed similarity matrix may serve as an input to various classifiers, which predict the synchronization offset.}
    \label{fig:videosync-overview}
\end{figure*}

In this Section, we describe our proposed approach to building a video synchronization AI system, \textbf{VideoSync}. Its overview is illustrated in~\cref{fig:videosync-overview}. Given two video files capturing the same scene from different perspectives as input, the system’s objective is to determine the integer synchronization offset between them. The inference pipeline consists of three stages. It begins with passing each of the input videos through an embedding model. The embedding model generates an embedding vector for every frame within each video. We use the CARL~\citep{SCL} embedding model for our main experiments, while InternVideo2~\citep{internvideo2} serves as a baseline alternative. Other models, such as V-JEPA~\citep{V-JEPA}, VideoMAE~\citep{videomae}, and ViClip~\citep{internvid}, were considered but did not yield meaningfully different results compared to InternVideo2.

In the next stage, we compute the matrix of pairwise negative Euclidean distance (as done by SeSyn-Net in~\cite{HumanPose}) between the embedding vectors corresponding to frames in different videos. Each entry in this matrix represents the similarity between a pair of frames belonging to the two videos. 

Finally, the softmaxed similarity matrix is used as an input to a classifier, which outputs an integer label that can be interpreted as the prediction of synchronization offset. We consider several classifiers, both hand-crafted and learned:

\begin{itemize}
\item \textbf{Argmax Method}: This is a manually constructed classifier used by SeSyn-Net. We compute the row-wise argmax of the similarity matrix, subtract the row indices from the result, and return the median value of the resulting vector.
\item \textbf{Dynamic Time Warping (DTW)}: Taken from video alignment methods, this approach is similar to the argmax approach, with DTW replacing the argmax operation. We apply the DTW algorithm~\citep{dtw} to the similarity matrix and compute the median of the offsets obtained by subtracting the row indices from the DTW results.
\item \textbf{Logistic Regression}: This is a trained classifier. Pad the similarity matrix to a size of 256x256 and train a multiclass logistic regression model to predict synchronization offsets in the $[-30, +30]$ range.
\item \textbf{Support Vector Machine (SVM)}: We employ scikit-learn's \texttt{SVR} to predict synchronization offsets as a regression task. The similarity matrix is used as input, and predictions are rounded to the nearest integer within the range $[-30, +30]$.
\item \textbf{Multi-Layer Perceptron (MLP)}: We use scikit-learn's \texttt{MLPClassifier} with 3 hidden layers using the padded similarity matrix as input. The MLP model predicts synchronization offsets as discrete classes within the range $[-30, +30]$.
\item \textbf{Convolutional Neural Network (CNN)}: We design a CNN using PyTorch with four convolutional layers followed by a classifier head. The network is trained using CrossEntropyLoss and optimized with AdamW to predict synchronization offsets as classification labels in the range $[-30, +30]$.
\item \textbf{Naive guess (baseline)}: As a simple baseline, we present a naive guess algorithm that always guesses 0 as the sync offset. Because our input video pairs' label sync offsets are uniformly distributed within range $[-30, 30]$, the expected value of the absolute frame error for this strategy is 15.
\end{itemize}

We describe them in more detail in Section 3 of supplementary material.

We extend the methodology proposed by SeSyn-Net~\citep{HumanPose} by removing its dependence on domain-specific feature extraction mechanisms, such as human pose estimation. Instead, we introduce the flexibility to substitute the embedding model with any generic video embedding model, broadening its applicability. 

\section{Implementation Details}
For our experiments, we used PyTorch 1.10 and used 4 * NVIDIA V100 GPUs for pre-training. We also used RTX 3090 for various evaluation tasks. For reproducibility, we place the datasets used for evaluation along with all of the preprocessing, training, and evaluation code in open repositories on GitHub.\footnote{Link to anonymized codebase: https://github.com/VideoSyncAI/videosync}

\subsection{Datasets}

We used video pairs derived from three datasets: NTU RGB+D\citep{NTUDataset}, Pouring\citep{PouringDataset}, and CMU Panoptic~\citep{CMUDataset} for our experiments.

NTU RGB+D is a large-scale dataset of human activities captured from multiple viewpoints using depth sensors, making it well-suited for evaluating synchronization across diverse perspectives. For this dataset, we followed the configuration outlined by \cite{HumanPose}, where the videos have an average length of 120 frames. The dataset includes 7,524 videos for training and 3,596 videos for testing. We modified the video lengths to ensure both videos in the pair had the same duration, as discussed in~\cref{sec:preprocessing}.

The Pouring dataset~\citep{PouringDataset} consists of video pairs captured from different angles, focusing on close-up shots of various liquids being poured between containers. This dataset is particularly useful for evaluating synchronization in fine-grained, repetitive motion scenarios. It consists of 266, 34, and 84 videos in the training, validation, and testing sets, respectively.

The CMU Panoptic dataset\citep{CMUDataset} is a multi-view dataset featuring large-scale motion capture recordings of human activities. From this dataset, we constructed two distinct subsets: CMU Pose, which focuses on single-person actions, and CMU Multi Human, which includes interactions among multiple people. We detail the dataset construction process in Section 2 of supplementary material. After pre-processing, we obtained 5,156 videos (3,598 training and 1,558 testing) for the CMU Pose dataset and 2,236 videos (1,554 training and 682 testing) for the CMU Multi Human dataset. All processed videos have a frame rate of 29.97 fps and a duration of 8 seconds.

\subsection{Preprocessing}
\label{sec:preprocessing}
Our task at hand is synchronizing two video recordings of the same scene taken from different angles with unknown time offsets. To construct relevant training and evaluation datasets, we took the datasets of pre-synchronized video recordings and injected temporal offsets sampled uniformly from the interval of $\pm 30$ frames.

\begin{figure}
    \centering
    \includegraphics[width=0.9\linewidth]{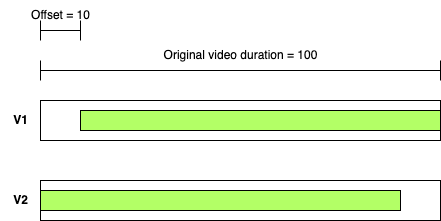}
    \caption{Illustrating our approach to keep both videos within a pair with identical length after temporal cropping. In this example, we inject an offset of 10 frames to V1. In order to ensure both processed videos have the same duration, we also crop 10 frames off the end of the second video V2.}
    \label{fig:video-durations}
\end{figure}

Our analysis in~\cref{sec:sesyn-net-reanalysis} shows that injecting the offset into only one of the videos can lead to inadequate experimental results due to information leaks through the difference in video duration. Thus, we applied a technique demonstrated in~\cref{fig:video-durations} to crop both videos in a pair and ensure their durations are identical.

For all datasets, we resized (by stretching) the frame dimension to 224x224 pixels to align with the format originally used by CARL.

\subsection{Embedding model architecture}

This section details the architecture of the main video embedding model used in our experiments. We adopted the Contrastive Action Representation Learning (CARL) model trained by \citet{SCL}, which leverages Sequence Contrastive Loss (SCL) to extract representative frame-level embeddings. The model’s architecture is illustrated in Figure 3 of \citet{SCL}. We utilized two sets of pre-trained weights:

\begin{itemize}
\item \textbf{NTU Pre-trained Weights}: We trained the CARL embedding model on the NTU dataset for 194 epochs from scratch, using pre-trained ResNet-50~\citep{resnet} backbone as it was done by~\cite{SCL}.

For model training, we followed \cite{SCL} by using only the first video from each pair during pre-training, as the model does not require video pairs in this phase. We employed a cosine annealing learning rate schedule with an initial learning rate of $10^{-4}$, maintaining a fixed rate within each epoch. Following the original configuration, we set NUM\_FRAMES=80. These implementation details highlight the importance of careful model tuning and pre-training setup for optimal performance across diverse datasets.

\item \textbf{K400 Pre-trained Weights}: Provided by \cite{SCL}, these weights resulted from training on the Kinetics-400 (K400) dataset \citep{K400Dataset} for 29 epochs.
\end{itemize}

We also employed the InternVideo2 model~\citep{internvideo2} as an alternative to the CARL embedding model. Specifically, we utilized the pre-trained weights from "\verb|InternVideo2-stage2_1b-224p-f4|" which processes input sequences of 4 frames at a resolution of 224x224 pixels and generates a 512-dimensional embedding vector. To obtain per-frame embeddings, we applied a sliding window approach with a stride of one frame over the video frames.

\subsection{Metric}

To evaluate the performance of our video synchronization method, we adopted absolute frame error as the primary metric, following the approach in \cite{HumanPose}. This choice facilitates direct comparison with their results and serves as an intuitive measure of synchronization accuracy.
\section{Experimental Results}\label{sec:exps}

\subsection{Exploring Vision-Language Models for Synchronization}

Before developing our specialized synchronization system, we investigated the potential of using state-of-the-art Vision-Language Models (VLMs) as end-to-end synchronization predictors. Given the recent success of VLMs in various vision-language tasks \citep{gpt-4o}, we hypothesized that these models might be capable of directly inferring temporal relationships between video pairs.

We conducted experiments using OpenAI's gpt-4o-mini model on a subset of 50 video pairs from the NTU test dataset. For each video, we extracted 30 frames using a downsampling ratio of 4:1, which allowed us to cover the full temporal range while staying within the model's input constraints. The model was prompted to predict the frame offset between each video pair, with clear instructions (shown in Listing 1 of supplementary material) about the expected output format and the interpretation of positive and negative offsets.

Our results showed that the VLM approach performed no better than naive guessing, with a mean absolute frame error of approximately 15 frames. Analysis of the predicted offsets revealed two key limitations:

\begin{itemize}
\item The model showed a strong bias toward predicting zero offset, despite the ground truth offsets following a roughly uniform distribution between -30 and 30 frames.

\item When the model did predict non-zero offsets, it showed an unexplained bias toward negative values, suggesting potential difficulties in establishing consistent temporal ordering between video pairs.

\end{itemize}

We initially attempted to use larger and more powerful "gpt-4o" for this experiment but encountered practical limitations. The model rejected requests exceeding 18 frames per video and imposed strict rate limits that prevented processing multiple video pairs in high enough temporal resolutions. In contrast, "gpt-4o-mini" allowed for processing more frames (Up to 30 per video) and had more lenient rate limits, making it the model of choice for this experiment. Both models frequently outputted instructions on how to write code to synchronize two videos, rather than directly predicting the sync offset. To mitigate this, we used the "response\_format" parameter to enforce a fixed JSON output format. Additionally, we tested Google's "gemini-2.0-flash-thinking-exp-01-21" model, but the results were similar.

These findings underscore the challenges of using general-purpose VLMs for specialized temporal reasoning tasks. While VLMs excel at semantic understanding and cross-modal reasoning, they appear to struggle with precise temporal alignment tasks that require fine-grained motion understanding. This insight motivated the development of our specialized synchronization system, which is discussed in the following sections.

\subsection{SeSyn-Net Reanalysis}
\label{sec:sesyn-net-reanalysis}
In our experiments, we revisited the original SeSyn-Net setup and observed an unusual phenomenon: replacing one video's input embeddings with random noise while preserving the shape of the similarity matrix resulted in better evaluation performance than the results reported in the original paper. This discovery raised questions about the robustness and validity of SeSyn-Net's methodology for calculating synchronization offsets.

\subsubsection{Original Setup and Reproduced Results}
In the SeSyn-Net framework, the authors use human pose data to extract frame embeddings, construct a similarity matrix between two videos, and apply the "argmax" algorithm to determine synchronization offsets. We successfully reproduced the reported results using the provided pre-trained weights and benchmark datasets, validating that the method achieves high accuracy for the original input embeddings.

\subsubsection{Random Noise Substitution}
To explore the robustness of the approach, we replaced the embeddings of one video (note that in \cite{HumanPose}, the video frames are pre-processed, and it uses human joint data as input) with random noise generated using \texttt{torch.randn\_like()}, which produces normally distributed random values, and then scaled to match the original data's mean and variance. The input tensor maintained the same shape as the original embeddings, i.e., \texttt{[1, 3, <num\_time\_steps>, 17, 1]}, where the dimensions other than \texttt{num\_time\_steps} correspond to human joint positions. Importantly, the duration of the videos (i.e., \texttt{num\_time\_steps}) was preserved to ensure the shape of the similarity matrix remained consistent with the original setup.

We evaluated the model using similarity matrices constructed from noise embeddings compared to the original embeddings. The results revealed an unexpected anomaly: not only did the metric remain high, but the performance improved when using noise embeddings. This improvement was attributed to increased randomness in the middle portion of the similarity matrix (Figure 2 of supplementary material), which caused the "argmax" algorithm to favor the correct offset more frequently.

\subsubsection{Methodological Implications}
The root cause of this anomaly lies in two key biases in SeSyn-Net's design:
\begin{enumerate}
\item \textbf{Preprocessing Bias}: The preprocessing step consistently cropped the beginning of the videos and synchronized the ends, introducing positional alignment artifacts into the similarity matrix.
\item \textbf{Positional Embedding Bias}: The positional encoding in SeSyn-Net leads to a strong inherent similarity between frames that share the same positions. Thus, the embeddings at the beginning and end of the videos were highly similar, even when one video was replaced with random noise. This bias resulted in diagonal patterns in the similarity matrix that aligned well with the correct synchronization offset, as shown in Figure 2 of supplementary material.
\end{enumerate}

\subsubsection{Empirical Analysis}
To isolate the effect of the bias in the results that arise due to the video pre-processing, we tested SeSyn-Net under controlled conditions. The results are presented in~\cref{tab:sesyn-net-degradation}. 
\begin{itemize}
\item After applying the pre-processing technique illustrated in~\cref{fig:video-durations}, the durations of both videos in a pair are equal, and the model's performance degrades to the level of naive guessing, where the predicted offset is consistently equal to 0.
\item On the CMU dataset, which contains longer videos, the performance drop was less pronounced due to a reduced influence of the positional biases in the similarity matrix. However, when we cropped the CMU videos down to 120 frames to match the average duration of the videos in the NTU dataset, the model's performance again approached naive guessing.
\end{itemize}

These findings underscore significant weaknesses in SeSyn-Net's dependency on positional encoding and preprocessing biases. \textbf{The videos must be pre-processed using the scheme proposed in~\cref{fig:video-durations} for a fair comparison between algorithms.} Moving forward, we utilize the evaluation results for SeSyn-Net and other methods obtained on the benchmarks containing video pairs of equal length.

\begin{table}[h!]
    \centering
    \begin{tabular}{|p{2.3cm}|c|cp{1.1cm}|}
        \hline
        \multicolumn{1}{|p{2.3cm}|}{\centering \textbf{Pre-processing method}} & \textbf{NTU} & \textbf{CMU} & (frames) \\ \hline
        \multicolumn{1}{|p{2.3cm}|}{\centering \citep{HumanPose}} & $3.39 \pm 0.27$ & $2.27 \pm 0.62$ & (480)  \\ \hline
       \multirow{3}{=}{\centering \cref{fig:video-durations}} &\multirow{3}{*}{$15.45 \pm 0.41$ } & $4.83 \pm 1.01$  & (480) \\ \cline{3-4}
         & &$9.04 \pm 2.10$& (240)\\\cline{3-4}
         &  & $11.31 \pm 1.33$ & (120) \\ \hline
    \end{tabular}
    \caption{Degradation of SeSyn-Net's absolute frame error when using the same video duration for a video pair. Longer videos seem to improve the absolute frame error.}
    \label{tab:sesyn-net-degradation}
\end{table}

\subsection{Comparative Analysis of Synchronization Methods}
Since SeSyn-Net relies on human pose embeddings, its applicability is inherently limited to videos containing human subjects with clearly visible poses. To develop a more general-purpose video synchronization system, we replace human pose embeddings with general-purpose video embeddings. We consider two main categories of video representation models: (1) models designed for video alignment tasks, such as TCC, TCN, and CARL, with CARL being the most recent and performant while also not requiring multi-view training; and (2) ViT-based models, such as VideoMAE, V-JEPA, and InternVideo2, which have gained traction and been integrated into foundation vision-language models. While we evaluate both categories in our experiments, we use CARL as the primary embedding model due to its computational efficiency—it is over two orders of magnitude faster than InternVideo2. This speed advantage stems from architectural differences: CARL employs a hybrid structure combining a ResNet-50 backbone with three transformer layers, whereas InternVideo2 is a pure ViT model with 40 transformer layers, making it significantly more compute-intensive.

In Table \ref{tab:sesyn-net-vs-scl}, we compare the absolute frame error of SeSyn-Net and CARL across different synchronization methods. We evaluate several approaches including DTW, Argmax, and various machine learning models (logistic regression, SVM, MLP, and CNN). Since we utilize multiple embedding models as well as synchronization methods, we denote our system name with the following naming convention: \texttt{VideoSync-<embedding model>-<synchronization method>}. The results demonstrate that all methods outperform SeSyn-Net on the NTU dataset, with \texttt{VideoSync-CARL-CNN} achieving the best performance (8.72 ± 0.54 frames). The significant improvement over the baseline \texttt{VideoSync-CARL-Argmax} method (10.88 ± 0.47 frames) suggests that leveraging more sophisticated neural architectures can better capture the temporal relationships between video frames.

For the CMU dataset, a direct comparison with SeSyn-Net was not possible due to incompatibility of the provided data and lack of clear documentation about their file naming conventions and preprocessing pipeline. Instead, we created a comparable dataset using the same source (CMU Panoptic Dataset - Range of Motion \cite{CMUDataset}) to perform an indirect evaluation. In this setting, \texttt{VideoSync-CARL-CNN} (9.59 ± 0.67 frames) demonstrated competitive performance compared to SeSyn-Net (9.04 ± 2.10 frames), while outperforming the \texttt{VideoSync-CARL-Argmax} baseline (11.17 ± 0.76 frames). Notably, simpler models like \texttt{VideoSync-CARL-LogReg} and \texttt{VideoSync-CARL-MLP} showed degraded performance on the CMU dataset (13.82 ± 0.90 and 21.38 ± 1.20 frames, respectively), likely due to their limited capacity to handle the increased complexity and longer durations present in this dataset.

The superior performance of CNN across both datasets can be attributed to its ability to learn hierarchical representations and capture both local and global temporal patterns. In contrast, simpler models like logistic regression and MLP, while performing well on the more controlled NTU dataset, showed limitations when dealing with the more diverse CMU dataset. This suggests that architectural complexity plays a crucial role in robust synchronization across different domains.



\begin{table}[h!]
    \centering
    \begin{tabular}{|c|c|c|}
        \hline
        \textbf{Method} & \textbf{NTU}        & \textbf{CMU}       \\ \hline
        SeSyn-Net       & $15.45 \pm 0.41$    & $\mathbf{9.04 \pm 2.10}$    \\ \hline
        VideoSync-CARL-Argmax   & $10.88 \pm 0.47$    & $11.17 \pm 0.76$   \\ \hline
        VideoSync-CARL-DTW       & $11.33 \pm 0.49$    & $10.90 \pm 0.76$   \\ \hline
        VideoSync-CARL-LogReg   & $9.85 \pm 0.58$     & $13.82 \pm 0.90$   \\ \hline
        VideoSync-CARL-SVM       & $10.56 \pm 0.42$    & $10.62 \pm 0.54$   \\ \hline
        VideoSync-CARL-MLP       & $\underline{9.20 \pm 0.55}$     & $21.38 \pm 1.20$   \\ \hline
        VideoSync-CARL-CNN       & $\mathbf{8.72 \pm 0.54 }$    & $\underline{9.59 \pm 0.67}$    \\ \hline      

    \end{tabular}
    \caption{SeSyn-Net vs. VideoSync-CARL: We measure absolute frame errors between SeSyn-Net and VideoSync-CARL using the same data pre-processing strategy. We make an indirect comparison for the CMU dataset as we could not reproduce SeSyn-Net's data pre-processing. \textbf{Bold} indicates the best result, and \underline{underline} indicates the second-best result.}
    \label{tab:sesyn-net-vs-scl}
\end{table}

\subsection{Effect of Video Duration on Synchronization Performance}
\label{sec:effect-of-video-duration}

In Section~\ref{sec:sesyn-net-reanalysis}, we observed that the SeSyn-Net model performed better on the CMU dataset. We attribute this result to differences in video durations between datasets. In this section, we investigate a similar question for the VideoSync-CARL model: Does the VideoSync-CARL method exhibit sensitivity to video duration? To explore this, we evaluated the VideoSync-CARL model, pre-trained on the K400 dataset, on a subset of the CMU Pose dataset comprising 182 video pairs. Figure 3 of supplementary material presents the results, where the absolute frame error was measured for video durations ranging from 80 to 480 frames to assess the effect of video length on performance.

Our analysis revealed that the absolute frame error decreased as video duration increased from 80 to 240 frames. This improvement can be attributed to the greater amount of information available in longer videos, which facilitates more reliable synchronization by providing additional frames for comparison and alignment.

Interestingly, beyond 240 frames, the absolute frame error reached a plateau and did not further decrease with increasing video duration. One plausible explanation is a distribution shift in the input data. Longer video durations may deviate significantly from the distribution of the data used for CARL training, which primarily consisted of videos approximately 80 frames long. This distribution mismatch may challenge the model's ability to maintain accuracy, preventing it from taking advantage of the additional information contained in much longer videos.

These findings underscore the importance of training models on datasets that reflect the characteristics of the target input conditions. When applying the CARL model to substantially longer videos, retraining or fine-tuning on an appropriately curated dataset may be necessary to ensure consistent synchronization performance.

\subsection{Synchronization of Other Datasets}

In the previous work, SeSyn-Net focused on automated synchronization of videos capturing single-human activities. The multi-human synchronization was achieved in a multi-step process by first extracting human pose embeddings of each person appearing on the screen and then averaging the embeddings. We argue that this approach is not practical and that a method that relies on human pose is not generalizable in videos that don't contain human subjects at all. In this section, we describe the evaluation of the \texttt{VideoSync-CARL-Argmax} method for the capability of synchronizing videos capturing scenes outside typical single-human activities. We utilize two datasets for this experiment: The pouring dataset and the CMU Multi Human dataset. The pouring dataset consists of pairs of videos shot at different angles that are close-up shots of various liquids being poured from one container to another. We construct the CMU Multi Human dataset by taking various datasets (as shown in Section 2 of supplementary material) from the CMU Panoptic Dataset, where multiple human subjects appear at the same time.
In Table~\ref{tab:other-datasets-synchronization}, we present the synchronization performance for the above-mentioned datasets. For the CMU Multi Human Dataset, \texttt{VideoSync-CARL-Argmax} performs slightly worse than the naive algorithm that always guesses 0 as the synchronization offset. This shows the challenges of synchronizing videos capturing multiple humans. When we inspected individual videos, we noticed the movements were often small, and it seemed hard even for humans to synchronize manually. In the case of the Pouring dataset, \texttt{VideoSync-CARL-Argmax} performed at 11.14 frames, which is much better than the naive guess. Note that SeSyn-Net is not feasible for this dataset as the Pouring dataset does not contain human subjects in the videos. We also note that the K400 pre-trained model performs much better than the NTU pre-trained model. This is likely because K400 dataset is much more diverse, larger in size (~250k vs. ~4k in dataset size), and likely to contain non-human videos. The error bound for the Pouring dataset is large because we only had access to 84 videos in the dataset.

\begin{table*}[h!]
    \centering
    \begin{tabular}{|c|c|c|c|}
        \hline
        \textbf{Method}       & \textbf{CMU Multi Human} & \textbf{Pouring} \\ \hline
        VideoSync-CARL-Argmax (NTU Pretrained)  & $17.99 \pm 1.40$         & $21.57 \pm 5.02$ \\ \hline
        VideoSync-CARL-Argmax (K400 Pretrained) & $ \mathbf{16.50 \pm 1.25}$         & $\mathbf{11.14 \pm 3.96}$ \\ \hline
    \end{tabular}
    \caption{Absolute frame error of VideoSync-CARL-argmax for datasets that contain no human or more than one human subjects. \textbf{Bold} indicates the best result.}
    \label{tab:other-datasets-synchronization}
\end{table*}

\subsection{Synchronization Using an Alternative Model}

As an alternative to CARL framewise embedding model, we evaluated synchronization performance using the InternVideo2 model~\citep{internvideo2}, selected for its strong performance in video-related tasks such as Temporal Action Localization and Video Retrieval. InternVideo2 is a ViT-based model, sharing architectural similarities with other foundational video models like VideoMAE and V-JEPA. In our experiments, we computed embeddings for sliding windows of 4 consecutive frames, treating each embedding as a per-frame representation. The synchronization offsets between videos were determined using the Argmax algorithm described in Section 3.1 of supplementary material.

Table~\ref{tab:sync-results-internvideo2} compares the synchronization performance of InternVideo2 paired with Argmax synchronization method (\texttt{VideoSync-InternVideo2-Argmax}) with \texttt{VideoSync-CARL-Argmax} baseline. For the adjusted InternVideo2 (denoted as InternVideo2 (adj.)), sync offsets exceeding $\pm30$ frames were set to 0. This adjustment compensates for InternVideo2's tendency to predict extreme offsets outside the $\pm30$ range. This behavior arises because InternVideo2 computes embeddings for 4-frame sliding windows, without leveraging global positional encodings that differentiate frames across the entire video. Without these positional cues, InternVideo2 can predict widely varying offsets, especially in challenging scenarios. The adjustment scheme mitigates this issue by suppressing predictions beyond $\pm30$ frames, effectively penalizing extreme offsets.

In contrast, CARL (and SeSyn-Net) exhibited a bias toward predicting offsets near zero in challenging scenarios, as its global positional encodings span the entire video. This often unfairly advantages CARL when faced with difficult synchronization tasks. For instance, Figure 4(a) of supplementary material highlights this bias, showing strong diagonal dominance in CARL's similarity matrix even when one video is replaced with random noise.

From Table~\ref{tab:sync-results-internvideo2}, CARL outperforms the raw InternVideo2 results for the CMU Pose and CMU Multi Human datasets. However, with the adjustment scheme applied, InternVideo2 performs comparably to CARL. Notably, on the Pouring dataset, InternVideo2 (both adjusted and non-adjusted) significantly outperforms CARL. This discrepancy likely stems from differences in pretraining datasets: CARL was pretrained on K400, which emphasizes human actions, while InternVideo2 leveraged a more diverse dataset. Since the Pouring dataset primarily focuses on liquid-pouring actions with minimal human-centric content, InternVideo2's superior performance aligns with its broader pretraining.

Figure 5 of supplementary material illustrates examples of good and bad synchronization using InternVideo2. Successful synchronization often corresponds to scenes with significant motion, as evident in clear diagonal alignments in the similarity matrix (Figure 5(a) of supplementary material). In contrast, failures frequently occur in scenes with minimal movement, leading to vertical patterns in the similarity matrix where many frames from one video align with a single frame of the other (Figure 5(b) of supplementary material). Addressing this limitation could be a focus for future work.


\begin{table*}[h!]
\centering
\begin{tabular}{|l|c|c|c|c|}
\hline
\textbf{Method}                      & \textbf{NTU}               & \textbf{CMU Pose}        & \textbf{CMU Multi Human}     & \textbf{Pouring}      \\ \hline
VideoSync-CARL-Argmax                & $13.63\pm0.43          $   & $\mathbf{11.17\pm0.76}$  & $\underline{16.50\pm1.25}$   & $11.14\pm3.96        $ \\ \hline
VideoSync-InternVideo2-Argmax        & $\underline{13.39\pm1.98}$ & $18.76\pm2.77         $  & $38.09\pm4.72         $      & $\mathbf{8.86\pm3.34}$ \\ \hline
VideoSync-InternVideo2-Argmax (adj.) & $\mathbf{11.33\pm1.62} $   &$\underline{12.64\pm1.48}$& $\mathbf{16.03\pm1.45}$      & $\underline{8.97\pm3.34}$ \\ \hline
\end{tabular}
\caption{Comparison of synchronization results between CARL and InternVideo2 embedding models. For InternVideo2 (adj.), we replaced sync offsets over $\pm30$ frames with 0. The \texttt{VideoSync-CARL-Argmax} method used K400 pre-trained weights for fair comparison. For InternVideo2 experiments, we used first 200 video pairs for NTU, CMU Pose, and CMU Multi Human datasets. \textbf{Bold} indicates the best result, and \underline{underline} indicates the second-best result.}
\label{tab:sync-results-internvideo2}
\end{table*}

\section{Future Work}\label{sec:future}

\subsection{End to end model}
While our best-performing approach—CARL for embedding extraction combined with a CNN for sync offset prediction (\texttt{VideoSync-CARL-CNN})—outperforms prior work and demonstrates strong generalizability, it remains constrained by the expressiveness of using a similarity matrix as input to the sync offset predictor. This limitation likely contributes to its weaker performance on more challenging datasets, such as those involving multiple humans.

A promising direction for future work is to develop an end-to-end model that directly takes two videos as input and predicts their synchronization offset. This approach could potentially provide a more specialized solution, leveraging richer representations and learning more complex temporal relationships, ultimately leading to improved performance.

\subsection{Long Video Synchronization}
Our current work focuses on short videos (8-16 seconds), but real-world applications often involve longer videos. Future research will have to address the computational challenges of embedding every frame and constructing similarity matrices for longer durations.

\subsection{Negative Sample Detection}
In realistic scenarios, not all video pairs can be synchronized. Our future work will include mechanisms to detect negative samples, enhancing the robustness of our synchronization approach.

\section{Conclusion}\label{sec:conc}
In this paper, we introduced \textbf{VideoSync}, a comprehensive approach to video synchronization that leverages general-purpose video representation models, moving beyond the constraints of traditional audio-based and human pose embedding methods. Our methodology employs a variety of embedding models and classifiers, demonstrating the flexibility and effectiveness of our approach across diverse datasets.

Our experiments highlight the potential of using models like CARL and InternVideo2 for video synchronization, with notable improvements achieved through advanced classifiers such as CNNs. These models not only outperform previous methods but also show promise in handling complex scenarios, such as multi-human interactions and non-human-centric videos.

The results underscore the importance of selecting appropriate pre-trained models and synchronization strategies tailored to the specific characteristics of the dataset. While CARL excels in scenarios with human actions, InternVideo2 shows superior performance in non-human-centric datasets, such as the Pouring dataset.

Looking forward, our future work will focus on developing end-to-end models that can directly predict synchronization offsets from video inputs, addressing the limitations of current methods that rely on similarity matrices. Additionally, we aim to extend our approach to handle longer video durations and incorporate mechanisms for detecting unsynchronizable video pairs, enhancing the robustness and applicability of our synchronization solutions in real-world scenarios.

Overall, our work lays the foundation for more versatile and scalable video synchronization techniques, paving the way for advancements in video analysis and processing applications.
{
    \small
    \bibliographystyle{ieeenat_fullname}
    \bibliography{main}
}

\end{document}